\documentclass[%
 reprint,
superscriptaddress,
 amsmath,amssymb,
 aps,
longbibliography
]{revtex4-2}

\usepackage{natbib}
\usepackage{graphicx}% Include figure files
\usepackage{dcolumn}% Align table columns on decimal point
\usepackage{bm}% bold math
\usepackage{url}
\usepackage[usenames,dvipsnames]{color}

\newcommand{\be}{\begin{equation}}
\newcommand{\ee}{\end{equation}}

\begin{document}

\title{Kraken: enabling joint trajectory prediction by utilizing Mode Transformer and
Greedy Mode Processing}

\author{Daniil S. Antonenko}
\author{Stepan Konev}
\author{Yuriy Biktairov}
\email{biktairo@usc.edu}
\author{Boris Yangel}
\affiliation{Yandex Self-Driving Group (as of October 2021)}

\begin{abstract}
Accurate and reliable motion prediction is essential for safe urban autonomy. 
The most prominent motion prediction approaches are based on modeling the distribution of possible future trajectories
of each actor in autonomous system's vicinity.
These "independent" marginal predictions might be accurate enough to properly describe casual driving situations where
the prediction target is not likely to interact with other actors.
They are, however, inadequate for modeling interactive situations where the actors' future trajectories are likely to intersect.
To mitigate this issue we propose Kraken --- a real-time trajectory prediction model capable of approximating pairwise interactions
between the actors as well as producing accurate marginal predictions.
Kraken relies on a simple Greedy Mode Processing technique allowing it to convert a factorized prediction for a pair of agents
into a physically-plausible joint prediction.
It also utilizes the Mode Transformer module to increase the diversity of predicted trajectories and make the joint
prediction more informative.
We evaluate Kraken on Waymo Motion Prediction challenge where it held the first place in the Interaction leaderboard and the
second place in the Motion leaderboard in October 2021.

\end{abstract}

\maketitle

%%%%%%%%%%%%%%%%
\section{Introduction}
\label{sec:Introduction}

%% [General intro] \\
Motion prediction in intense urban conditions is one of the few crucial problems in the modern self-driving technology \cite{ReviewParavarzar}.
Engineering a sufficiently reliable predictor is a challenging task, which has been extensively addressed over the last decade.
These attempts demonstrated a substantial progress in prediction quality and allowed to deploy autonomous driving in a number of metropolitan regions.
However, much remains to be done on the way to a safe full-self-driving capability without any human supervision. 
The key challenges include multi-modal nature of the possible future trajectories for each agent, social interactions between the agents, and location-to-location shifts in conditions, traffic patterns and trajectory distributions \cite{SHIFTS}.

Among the listed challenges a special place is taken by the inherently interactive nature of certain
safety-critical driving scenarios. When the intended trajectories of two select actors intersect
their motion can no longer be adequately modeled independently. Indeed, marginal predictions for two
cars approaching an unprotected intersection might include both cars either going through the
intersection or giving way to the other car. However, the resulting factorized prediction for the
joint motion of the cars then contains unlikely colliding trajectories and is, clearly, not a good
representation of the real joint distribution of future motion of the two actors.

A special and, possibly, the most important type of these driving scenarios involves a pair of
actors including the autonomous vehicle (AV) itself. These situations require special attention
from the planning subsystem of the AV to be resolved safely as they involve dangers of potential
collision. The inaccuracy of a factorized prediction then becomes a major obstacle for the planning
subsystem which has to rely on various heuristics in order to compensate for it.

In this work we address the problem of producing adequate pairwise joint trajectory predictions
and introduce a novel motion prediction model named Kraken. Kraken relies on a simple yet effective
Greedy Mode Processing technique to modify an initially factorized prediction for a pair of actors and
obtain a joint prediction complying with the physical non-collision constraints of the real world.
To enhance this conversion, Kraken also includes the Mode Transformer module, which facilitates accurate
modeling of interactions and, empirically, usually leads to more diverse predictions.

\section{Related work}

Contemporary motion prediction approaches utilize a variety of deep neural architectures,
including convolution networks (CNN) and transformers.
The latter have become incredibly popular in machine learning in the last few years and
were shown to be a useful tool in the motion prediction \cite{SceneTransformer}.
In this work, we use transformer-inspired ideas throughout the model's architecture and
show that it significantly increases the quality of the predictions.

With respect to input format used for the representation of the scene, one can divide approaches used
in motion prediction in two classes:
\begin{enumerate}
    \item ones that use rasterization to prepare bird-eye
    view renderings of the environment as a raster image to then be used as an input for
    CNN \cite{chai2019multipath, cui2019multimodal, casas2018intentnet, hong2019rules,
    KonevWaymo, PRANK};
    \item "vector" models that use architectures that can directly process numerical
    representations of the scene \cite{VectorNet, liang2020learning, casas2020spagnn}.
\end{enumerate}
The second class of models has been showing a lot of promise lately.
Particularly, they require lower computational resources. Nonetheless, we claim that CNN-based
models are still potent in practice. In this work, we present a model relying on a
rasterized scene representation and demonstrate that it is competitive compared to the state-of-the-art solutions.

Development of motion prediction via deep neural networks is impossible without a high-quality dataset,
which should have enough size and diversity to adjust the predictor to fit miscellaneous road scenarios.
Fortunately, there is a number of publicly available datasets, which are commonly used in the
community \cite{Argoverse, WaymoDataset, SHIFTS}.
In this work, we make use of Waymo Open Motion Dataset \cite{WaymoDataset} for the experimental
evaluation since it's one of the largest-scale datasets available and contains rich annotations of
the road structure. We evaluate our approach according to the format of Waymo Motion and Interaction
Prediction Challenges 2021 \cite{WaymoMotionPrediction}. That allows to compare the
performance of our model to other solutions submitted to the public leaderboard.

A quick summary of the contributions we make in this paper is:
\begin{itemize}
    \item we introduce the Greedy Mode Processing technique which enables one to turn a factorized
    motion prediction of a pair of actors into a feasible joint prediction;
    \item we introduce the Mode Transformer module which noticeably increases the quality of
    multimodal prediction;
    \item we evaluate our motion prediction model --- Kraken --- on Waymo Open Motion Dataset and
    confirm its effectiveness by finding that it places the first on the Interaction leaderboard
    and the second on the Motion leaderboard of the Waymo Motion prediction Challenge (as of October 2021).
\end{itemize}

The rest of the paper is organized as follows: in Section~\ref{sec:methods} we describe data
preprocessing and detailed structure of our model. We emphasize the role of Mode Transformer module
(Section~\ref{sec:decoder}) and Greedy Mode Processing (Section~\ref{sec:NMS}).
The evaluation of the results and ablation studies are presented in Section~\ref{sec:results}. The conclusion is given in Section~\ref{sec:conclusion}.

%%%%%%%%%%%%%%%%
\section{Kraken}
\label{sec:methods}

We now introduce Kraken and describe its key components in details. We also specify the training
routine we follow and the way we choose appropriate values of hyperparameters.

\subsection{Probabilistic model}

\begin{figure*}
    \includegraphics[width=0.85\linewidth]{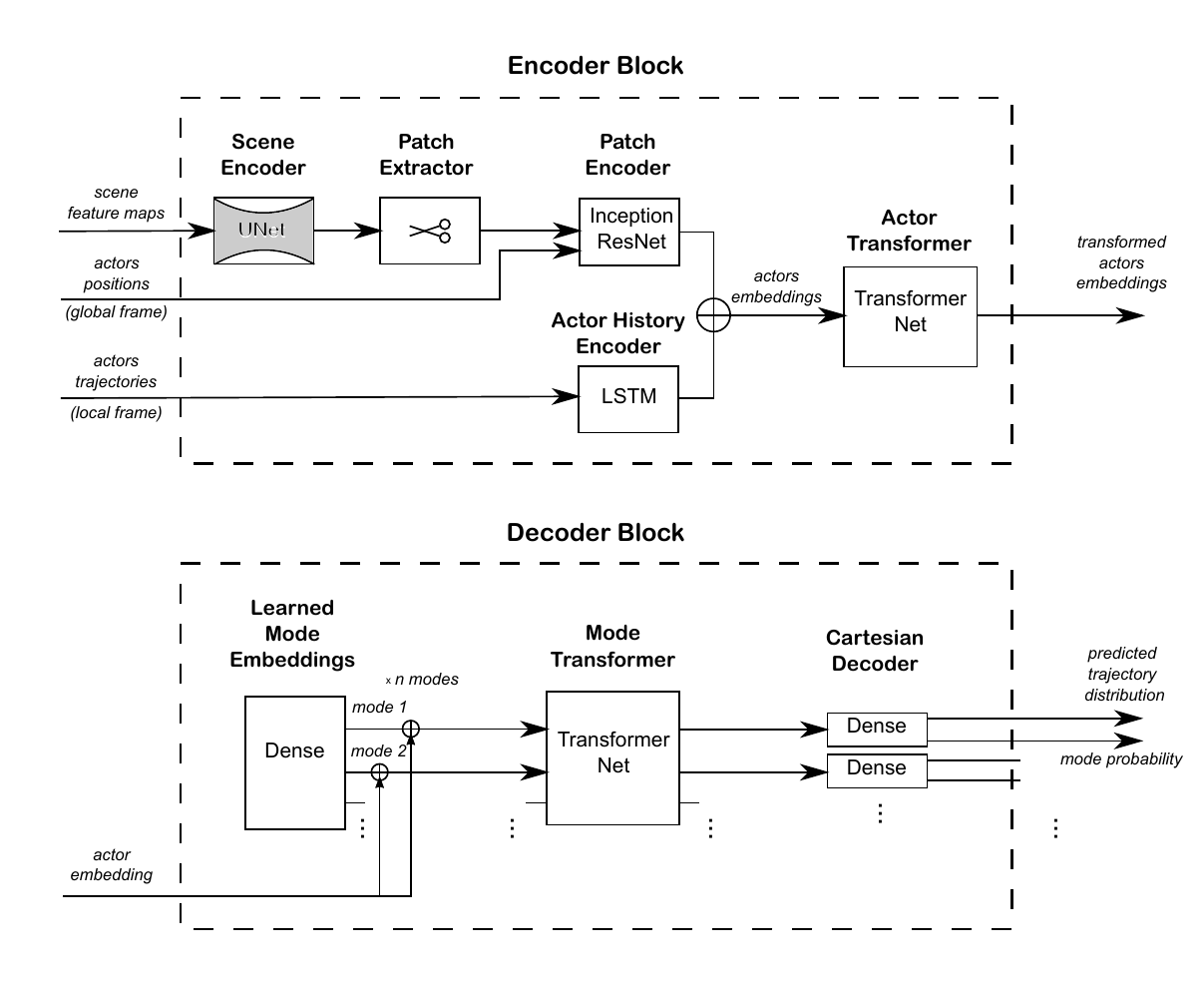}
    \caption{Architecture of the model represented as encoder and decoder blocks.}
    \label{fig:scheme}
\end{figure*}

We use a probabilistic model that represents the distribution of possible future trajectories
as a mixture of six unimodal components (``modes'') with probabilities $p_{\alpha}$.
We start with making independent predictions for each agent in the scene, so the probability of the future, where $i$-th agent moves along the trajectory $(x^{(i)}, y^{(i)})$ is expressed in the factorized form as:
\begin{equation}
\label{distribution_total}
    P(\{ x^{(i)}, y^{(i)}\} ) = \prod_i \sum_\alpha p_\alpha^{(i)} \cdot P_\alpha^{(i)} (x^{(i)}, y^{(i)}),
\end{equation}
where $\alpha$ enumerates the modes. 
Next, we choose the distribution for each mode to be factorized between the timesteps as well:
\begin{equation}
\label{distribution_exponential}
P_\alpha^{(i)}(x^{(i)}, y^{(i)}) \sim \prod_t \exp \left( - \frac{\left[(X_{\alpha t}^{(i)})^T \Sigma_{\alpha t}^{(i)} X_{\alpha t}^{(i)} \right]^{r/2}}{2} \right),
\end{equation}
where $\Sigma_{\alpha t}^{(i)}$ is the inverse covariance matrix and the displacements $X_{\alpha t}^{(i)}$ equal
\begin{equation}
 X_{\alpha t}^{(i)} = 
\begin{pmatrix}
x_t^{(i)} - x_{\alpha t}^{(i)} \vspace{0.15cm} \\ 
y_t^{(i)} - y_{\alpha t}^{(i)}
\end{pmatrix} .
\end{equation}
Here $x_{\alpha t}^{(i)}$ and $y_{\alpha t}^{(i)}$ are distribution parameters that coincide with
the expected value for the trajectory in mode $\alpha$.

The values of $x_{\alpha t}^{(i)}$ and $y_{\alpha t}^{(i)}$ should be predicted by the model, while $\Sigma_{\alpha t}^{(i)}$ can either be predicted or be set to a constant.
In the first case, the value of $\Sigma_{\alpha t}^{(i)}$ can provide additional information that
can be used in applications. For instance, the width of the predicted distribution can be used to
asses the certainty of the model prediction and help the planning subsystem of the AV choose an
appropriate behavior policy.
If one chooses to fix $\Sigma_{\alpha t}^{(i)} = \Sigma_{t}^{(i)}$, then it can be used as a tuning
knob in a likelihood-optimizing model to effectively attribute weights to actor types, prediction
times, etc. We explore both options in this work and discuss them in Sec.~\ref{sec:waymo_covariance}.

For the purposes of displacement-based evaluation we always use the expected values of the
coordinates \eqref{distribution_exponential} to generate predictions in form of concrete trajectories. 
Displacement power parameter $r$ is a hyperparameter and is tuned via cross-validation. 

We use the following parametrization for the inverse covariance matrix in Eq.~\eqref{distribution_exponential}:
\begin{equation}
    \Sigma = e^c \begin{pmatrix}
    e^a \cosh b & \sinh b \\
    \sinh b & e^{-a} \cosh b
    \end{pmatrix} .
\end{equation}
Thus, $\Sigma_{\alpha t}^{(i)}$ is parametrized by the three parameters $a_{\alpha t}^{(i)}$,
$b_{\alpha t}^{(i)}$, and $c_{\alpha t}^{(i)}$ accompanying the expected value parameters
$x_{\alpha t}^{(i)}$ and $y_{\alpha t}^{(i)}$. All in all, our model predicts five parameters
for a given agent ($i$), mode ($\alpha$), and timestamp ($t$).

As per common practice, we divide our model in encoding and decoding blocks (Fig.~\ref{fig:scheme}).
As mentioned earlier, we use a raster-based model, so the first stage is data preprocessing
(Sec.~\ref{sec:preprocessing}), which creates feature maps to be used as one of the inputs for
the encoding block. The other inputs are actor's past trajectory (represented in the
local coordinate frame) and global-frame actor position at the moment of prediction. 

CNN-based encoder (Sec.~\ref{sec:encoder}) uses both feature maps and actor history to
transform input into an embedding for each actor. 
Note that we use information on all actors in the scene, but evaluate embeddings only
for the selected ``interesting'' objects, which are flagged as potentially important for planning.
Next, for each actor independently, we generate six mode embeddings (Sec.~\ref{sec:decoder}).
Finally, each embedding is passed to the decoder (Sec.~\ref{sec:decoder}), which generates five parameters ($a_{\alpha t}^{(i)}$, $b_{\alpha t}^{(i)}$, $c_{\alpha t}^{(i)}$, $x_{\alpha t}^{(i)}$, $y_{\alpha t}^{(i)}$) for each of the six distribution components $P_{\alpha}^{(i)}$. Additionally, the decoder assesses the probability of each mode $p_{\alpha}^{(i)}$.  After proper normalization of the latter $p_{\alpha}^{(i)} \rightarrow p_{\alpha}^{(i)} / \sum_{\alpha} p_{\alpha}^{(i)}$, the total distribution is given by Eq.~ \eqref{distribution_total}.

We train three different decoders for each of the commonly encountered object types (vehicles,
pedestrians, and cyclists). The choice of the loss function is discussed in Sec.~\ref{sec:loss}.
In the following Sec.~\ref{sec:waymo_covariance} we discuss the effect of $\Sigma$-modeling
strategy on the diversity of the predictions.
The final stage is Greedy Mode Processing of the model predictions inspired by the
non-maximum suppression algorithm (NMS), which we describe in Sec.~\ref{sec:NMS}.

\subsection{Scene representation}
\label{sec:preprocessing}
Our model starts processing the data by preparing feature maps that represent the road and traffic
situation at the moment of prediction. We do not use the information on the traffic lights state
in the past timestamps. Feature maps are created in the form of a multi-channel $400 \times 400$
raster image of the scene bird-eye view, which is centered at the AV's position.
We use the resolution of $0.75 \text{m}$ per feature map pixel. Feature maps contain
the following channels:
\begin{itemize}
    \item lane centers, represented by curves in the binary format;
    \item lane speed limits, represented by a distinct number for vehicle and bicycle lanes;
    \item lane directions, represented by the tangent angle in the global frame;
    \item lane availability (binary), determined by the traffic lights signal, if any;
    \item crosswalks, represented by the binary curve, encircling the corresponding quadrangle;
    \item speed bumps, represented analogously to the previous channel;
    \item road polygons, represented by a binary mask;
    \item stop signs positions, binary mask;
    \item three separate channels for actors in the scene (vehicles, pedestrians, and cyclists),
    binary mask;
    \item six separate channels for in-plane components of vehicles, pedestrian, and cyclist's
    velocities, represented by real numbers.
\end{itemize}

Data preprocessor also prepares the history of actors positions in the local frame at the
frequency of $10 \text{Hz}$. If tracking information is absent for some of the timestamps,
we use a simple interpolation unless the gaps are too large.
The feature maps are prepared per scene rather than per actor.

\subsection{Encoder block}
\label{sec:encoder}

\paragraph{Feature maps encoding.}
We use a convolutional neural network to encode the feature maps. 
This procedure can be implemented in two ways \cite{SceneTransformer}: either one encodes a patch around each actor, which provides a more object-specific information in a pose-invariant way; or one encodes the whole scene, which saves resources and allows to encode global features in a unified manner.
We adopt a combined approach and firstly apply UNet to the scene-centric feature maps and then crop patches around each agent to be predicted and orient them according to the agent's pose. Generated patches are concatenated with additional channels, representing actors positions in the global coordinate frame and then passed to Inception ResNet network to generate the first part of the actor embedding. 

\paragraph{Actor history encoding.}
In parallel to the feature maps, our model encodes past actor trajectory via trainable LSTM network. Here we choose to represent trajectory in the local coordinate frame. Resulting actor history embedding is the second ingredient of the actor embedding.
It is then concatenated with the first part (obtained from the feature map) to form preliminary embedding for each actor to be predicted.

\paragraph{Actor transformer block.}
We then apply a TransformerNet to the actor embeddings to account for the social interactions. 
We use attention dimension of $256$ and Relu activation function, supplementing the transformer by a Dense layer. 
That important block finishes the encoding compartment of our model. 

\subsection{Decoder block and Mode Transformer}
\label{sec:decoder}

\paragraph{Generating mode embeddings.}
In the beginning of the decoder block, we promote each actor ($i$) embedding to six actor-mode embeddings parametrized by the indices ($i,\alpha$). This is done by using a Dense layer to create learned mode embeddings and concatenating them with the actor
embeddings.

\begin{figure}
    \centering
    \includegraphics[width=0.95\linewidth]{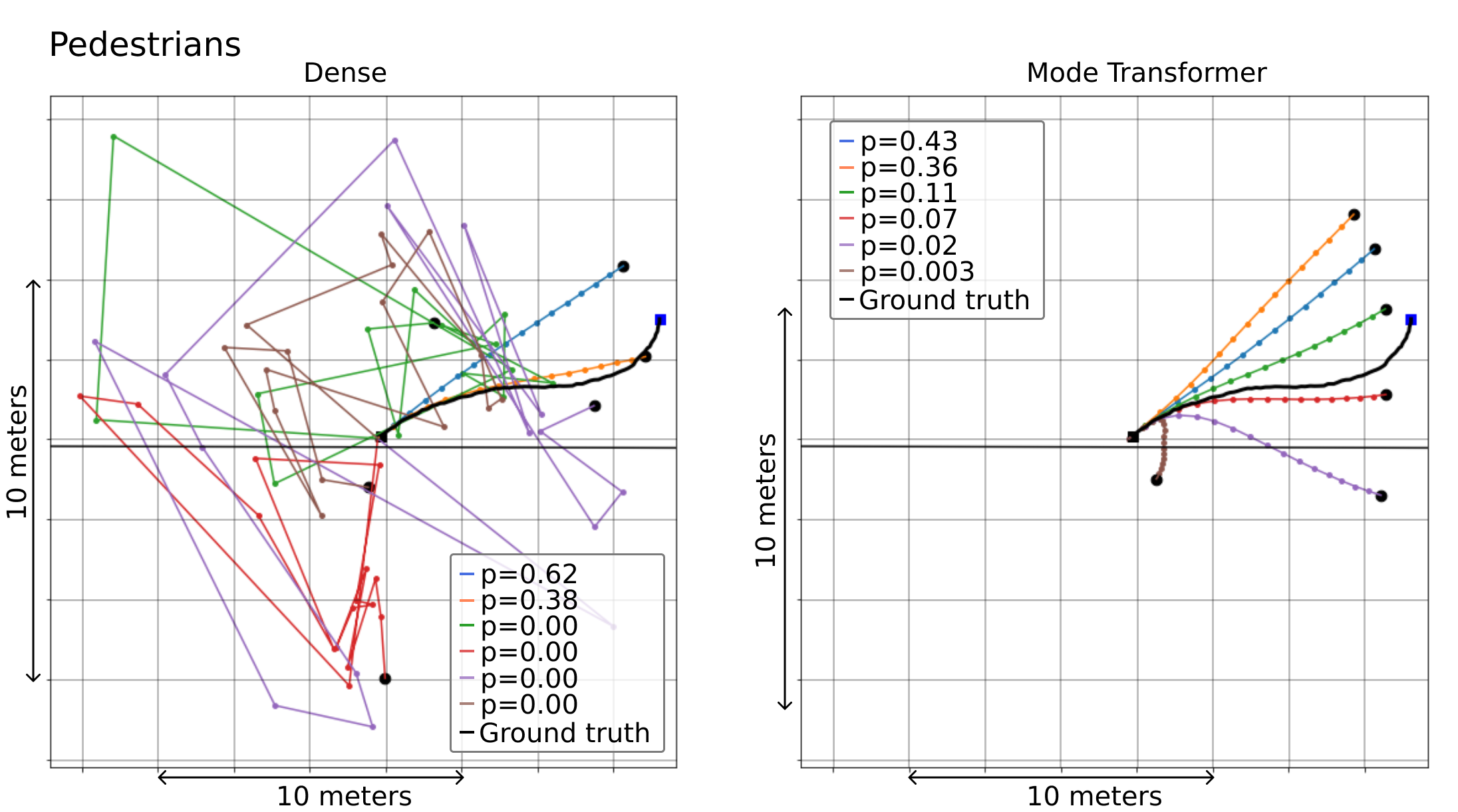}
    \includegraphics[width=0.95\linewidth]{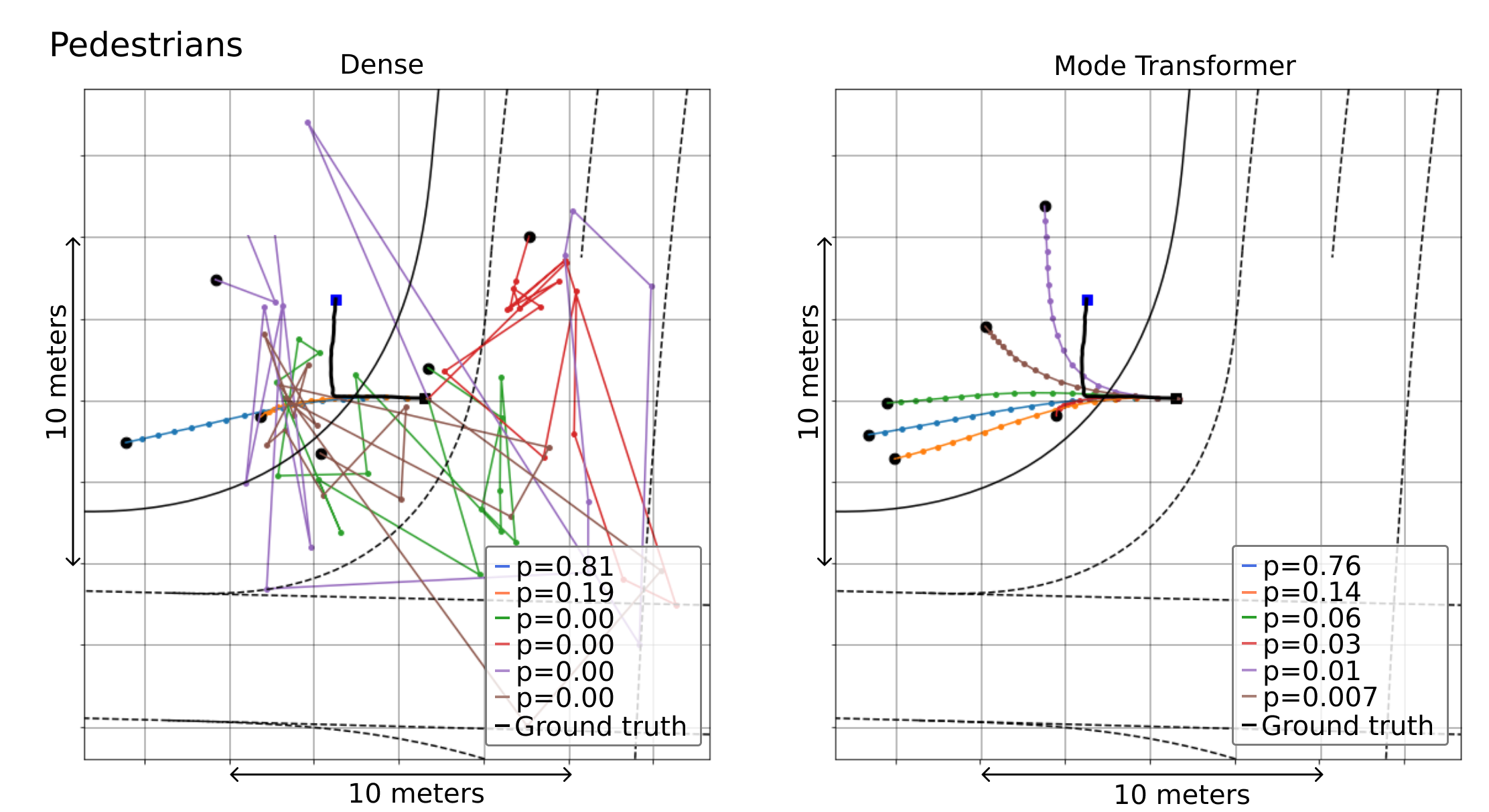}
    \caption{Examples of predictions with and without Mode Transformer block. Coordinates are given in meters.}
    \label{fig:MT_prediction}
\end{figure}

\paragraph{Mode transformer.}
This block constitutes one of the main developments of our work. 
Rather than processing each mode independently, we apply first the block that we call Mode Transformer.
It is implemented via a TransformerNet with attention dimension $256$ and followed by a Dense layer.
As we demonstrate in Sec.~\ref{sec:results}, introducing this block to our model significantly
increases the metrics. 
We interpret this quality increase as the effect of cooperation between the modes, which allows to
cover the distribution of possible future trajectories much more effectively.
In Fig. \ref{fig:MT_prediction} we provide examples of predicted trajectories to compare them
between model with Model Transformer and model with a Dense layer on its place. We find that
the predictions produced using the Mode Transformer are far less likely to contain noisy
modes with extremely low probabilities.

After processing with the Mode Transformer block, each mode embedding is separately put into
trajectory decoder, implemented as a dense network. It produces two outputs: mode probability
$p_{\alpha}^{(i)}$ for each agent ($i$) and mode ($\alpha$), and decoded trajectory distribution
at timestep $t$, represented by its expected values $x_{\alpha t}^{(i)}$, $y_{\alpha t}^{(i)}$
and covariance matrix $\Sigma_{\alpha t}^{(i)}$, which is parameterized by three variables
$a_{\alpha t}^{(i)}$, $b_{\alpha t}^{(i)}$, and $c_{\alpha t}^{(i)}$.
As we discuss below, we consider both predicted and fixed covariance matrices. 
In the latter case it not necessary to treat it as a decoder output. 

\subsection{Loss function and optimization}
\label{sec:loss}

We train the model by using the negative log-likelihood loss function obtained by substituting
ground truth trajectories $x_{\text{gt}}^{(i)}, y_{\text{gt}}^{(i)}$ into the distribution
\eqref{distribution_total}:
\begin{equation}
\label{loss}
    \mathcal{L} = - \sum_i \log \sum_\alpha p_\alpha^{(i)} \cdot P_\alpha^{(i)} (x_{\text{gt}}^{(i)}, y_{\text{gt}}^{(i)}) .
\end{equation}

We then train our model on $4-8$ NVIDIA A100 GPU devices by minimizing \eqref{loss} with Adam
optimizer and track the quality as evaluated on a held-out validation part of the dataset used. 
We use batch sizes of $152-300$ and validate the model multiple times per epoch due to the relatively
large size of the dataset. Finally, we use a learning rate scheduler that drops the learning rate
by a factor of $0.1$ whenever the loss does not improve by $10^{-3}$ in $10$ consecutive validations.

We compared training with the loss function, gathered from all actors in the scene with only
using the subset of objects marked as demonstrating interesting behavior in the dataset.
It appeared that the second option leads to the model, which shows a much better quality on the
validation dataset.
Also, inspired by some of the recent developments in the field, we experimented
with using the minADE metric directly as a part of the loss function. These experiments did not lead
to any noticeable improvement of the prediction's quality. 

\subsection{Increasing diversity of predictions}
\label{sec:waymo_covariance}

As mentioned above, one has to choose between the options of predicted or fixed $\Sigma$-matrix
in the distribution \eqref{distribution_exponential}. 
At this point it becomes important, which metric is being optimized because the trade-off behavior
between different metrics is observed.
In particular, we found that choosing a model-predicted covariance matrix leads to better values
for minADE metric while fixing it improves mAP metrics (see Sec.~\ref{sec:Waymo-dataset} for the definition), which corresponds to more diverse and
agent type aware predictions. 

Further enhancement of mAP can be achieved by adapting the distribution
\eqref{distribution_exponential} to actor- and timestamp- specific spacial tolerance levels as defined in the rules of Waymo Prediction challenge. 
We thus tune $t$-dependence of $\Sigma_t^{-1/2} = a + b t$, so that the standard deviation of
\eqref{distribution_exponential} linearly depends on time. We also make the distribution
anisotropic so that it is more stringent in the the direction, perpendicular to the ground truth
trajectory than along it. Finally, we introduce a dependence on the initial velocity,
which is supposed to reflect lower spacial accuracy expectations for the predictions of fast actors.
That only leaves unfixed the overall proportionality coefficient $\sigma$ scaling the matrix $\Sigma$.

\begin{figure}
    \includegraphics[width=0.99\linewidth]{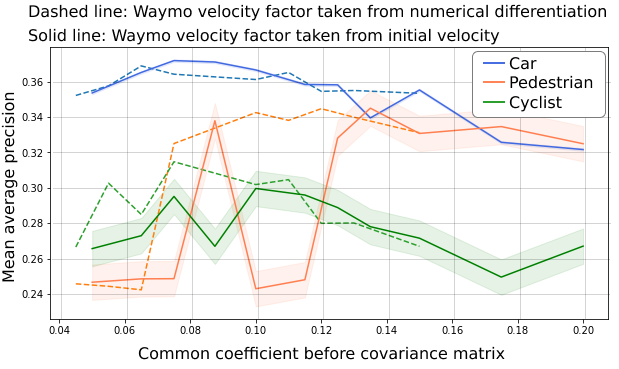}
    \caption{Optimizing the overall factor $\sigma$ for the covariance matrix, see Sec.~\ref{sec:waymo_covariance} for the definition. Different actor types peak at different values of $\sigma$.}
    \label{fig:covariance_matrix}
\end{figure}

We choose the value of this hyperparameter by optimizing mAP metric on the validation dataset (see
Fig.~\ref{fig:covariance_matrix}). We select a separate value for each actor type. 
Notably, the pedestrian and cyclist metrics showed rather big fluctuations leading to abrupt jumps
for pedestrian mAP in Fig.~\ref{fig:covariance_matrix}.

We also optimize the value of the `displacement power' parameter $r$ in
Eq.~\ref{distribution_exponential}.
The quality generally increases with the increase of $r$, but at $r > 1.5$ the training
process becomes rather unstable. Hence, we use the value of $r=1.5$ for all other experiments.
The comparison of different choices for $\Sigma$ are given in Tbl.~\ref{tbl:Ablations},
where `baseline' corresponds to the tuning described above, while `Constant Cov' is for
$\Sigma$ that does not depend on time and initial velocity (again, we tuned that constant value
for each actor type). 
Next, `Trainable Cov' is for $\Sigma_{\alpha t}^{(\alpha)}$ that should be predicted by the
model and `Tr. Cov w Min' represents the same configuration with a cutoff for $\Sigma$ that
forbids the standard deviation of \eqref{distribution_exponential} to drop below $3cm$.

%%%
\subsection{Greedy Mode Processing}
\label{sec:NMS}

In this section, we describe a method to boost the model performance with respect to the mAP metric of the Waymo Motion Prediction. We describe this metric in detail in Sec.~\ref{sec:Waymo-dataset}, while here we would like to emphasize that the it is essentially area under precision-recall curve, in which true positive predictions contribute more quality if they appear closer to the top of the list (sorted by predicted mode probabilities).
It is thus expected that one can improve mAP by filtering model predictions to eliminate `duplicates', which are predictions that lie too close to another prediction (compared to the scale of Waymo hit rate rules). 
Such filtering can be effectively implemented by a non-maximum suppression (NMS) algorithm (see e.g. Ref.~\cite{DenseTNT}).
\begin{figure}
    \includegraphics[width=0.99\linewidth]{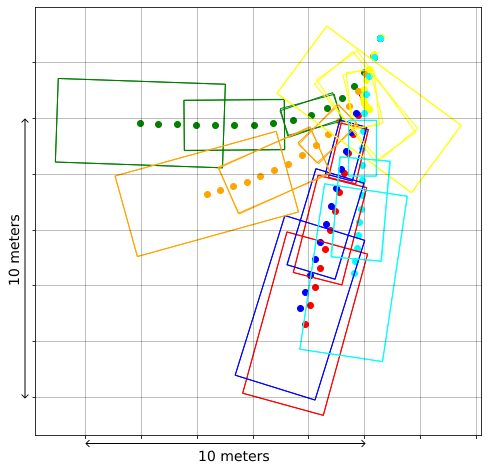}
    \caption{An example of a six-mode prediction for a single agent with Waymo hit
    windows shown at $3$, $5$, and $8$ s around each trajectory.
    One can clearly see that modes drawn in blue (2) and red (0) are too close to
    each other and one of them can be safely eliminated enhancing the mAP metric. The coordinates are given in meters.}
    \label{fig:NMS_example}
\end{figure}
In the Fig.~\ref{fig:NMS_example}, we show an example of a six-mode prediction for a single agent and indicate with rectangles hit windows of the mAP metric at $3$, $5$, and $8$-second timesteps aligning them along the predicted trajectories. 
One can easily observe that blue and red curves are almost identical, which means that if one of them is a hit, than the other is a hit as well with great probability. It is thus beneficial for the mAP to leave only one of the two predictions. 

Our implementation of the NMS algorithm consists of the following steps.
Firstly, for each agent we evaluate the $6\times6$ coincidence matrix of the modes. That requires selecting a specific criterion that would mark two predicted trajectories as coincident.
We tried several options for such criterion, including (i) comparing only the final points of the trajectories, (ii) comparing the
trajectories at $3$, $5$, and $8$ seconds with AND-logic afterwards as well as other similar choices.
We then picked the best strategy by optimizing mAP on the validation subset for each actor
type separately.

After the coincidence matrix is calculated, the next step is to eliminate redundant trajectories.
We start from the mode with the largest probability and then consider the bunch of modes that
are equivalent to it according to the matrix. One of the three elimination options is applied:
\begin{itemize}
    \item leave only the mode with the largest probability and suppress the other modes in the bunch;
    \item average object positions between the modes in the bunch at each timestep with equal weights;
    \item average as in the previous option, but use predicted mode probabilities as weights.
\end{itemize}
The resulting mode then aggregates the probabilities of the suppressed modes.
After a bunch is processed we exclude it from the consideration and apply the algorithm to the
rest of modes until all of them are exhausted. 
We chose one of the three options from the list above to maximize mAP value on the validation
dataset for each agent type separately.

The raw model and the model with NMS postprocessing are compared in Tbl.~\ref{tbl:Ablations}.
Applying NMS alogrithm turned out to provide a strong boost to the mAP score. 

\subsection{Pairwise joint prediction}
\label{sec:interaction}

We also apply the Greedy Mode Processing to generate feasible pairwise joint predictions given
the initial factorized prediction Kraken produces directly.
The factorized prediction processing consists of the following steps. 
For both objects in the pair we predict six possible future trajectories and estimated mode
probabilities as $p_\alpha^{(1)}$ and $p_\alpha^{(2)}$. Then, we compute $36$ probabilities
for all possible trajectory combinations assuming independent distribution of actors'
trajectories ($p_\alpha^{(1)} \cdot p_\beta^{(2)}$). We filter out the trajectory combinations
which lead to a "collision" according to a criterion similar to those introduced above.
Finally, we use the Greedy Mode Processing as described in the previous section, but apply it to the composite ``trajectory'', consisting from two trajectories for the two selected actors.
That allows to diversify the
final result and make the prediction more complete.

%%%%%%%%%%%%%%%%%%%%

\begin{table*}[t]
\caption{\label{tbl:motion_leaderboard} 
 Motion leaderboard of the Waymo Prediction Challenge (top-8).
}
\begin{ruledtabular}
\begin{tabular}{l|ccccc}
 Model &  minADE & minFDE  & Miss Rate & Overlap Rate & \textbf{mAP} \\
 \hline
multipath++\footnote{Public leaderboard contained two submissions for multipath++ model. In our table we include the best one (with respect to both mAP and minADE).} \cite{MultipathPlusPlus}
& 0.569 & 1.194 & 0.143 & 0.132 & 0.401 \\
\textbf{Kraken-NMS (our)} & 0.673 & 1.395 & 0.185 & 0.142 & 0.365 \\
Scene Transformer (M+NMS) \cite{SceneTransformer} & 0.678 & 1.376 & 0.198 & 0.146 & 0.337 \\
TVN & 0.750 & 1.584 & 0.183 & 0.145 & 0.334 \\
Tsinghua MARS - DenseTNT \cite{DenseTNT} & 1.039 & 1.551 & 0.157 & 0.178 & 0.328 \\
Star Platinum & 0.810 & 1.761 & 0.234 & 0.177 & 0.281 \\
ReCoAt \cite{ReCoAt} & 0.770 & 1.667 & 0.244 & 0.164 & 0.271 \\
AIR & 0.868 & 1.669 & 0.233 & 0.158 & 0.260 
\end{tabular}
\end{ruledtabular}
\end{table*}

\begin{table*}[t]
\caption{\label{tbl:interaction_leaderboard} 
Interaction leaderboard of the Waymo Prediction Challenge (top-7).
}
\begin{ruledtabular}
\begin{tabular}{l|ccccc}
 Model &  minADE & minFDE  & Miss Rate & Overlap Rate & \textbf{mAP} \\
 \hline
\textbf{Kraken-marginal (our)} & 1.037 & 2.406 & 0.512 & 0.214 & 0.165 \\
SceneTransformer(J) & 0.977 & 2.189 & 0.494 & 0.207 & 0.119 \\
AIR\^2 & 1.317 & 2.714 & 0.623 & 0.247 & 0.096 \\
King Crimson & 2.466 & 0.610 & 0.260 & 0.090 & – \\
HeatlRm4 & 1.420 & 3.260 & 0.722 & 0.284 & 0.084 \\
Waymo LSTM baseline & 1.906 & 5.028 & 0.775 & 0.341 & 0.052 \\
CNN-MultiRegressor & 2.585 & 6.370 & 0.891 & 0.308 & 0.034
\end{tabular}
\end{ruledtabular}
\end{table*}

\begin{table*}[t]
\caption{\label{tbl:Ablations} 
The results of the ablation study.}
\begin{ruledtabular}
\begin{tabular}{c|cccc|cccc}
 & mAP & mAP (Cars) & mAP (Peds) & mAP (Cycs) & minADE & minADE (Cars) & minADE (Peds) & minADE (Vehs) \\
 \hline
 baseline & 0.365 & 0.410 & 0.355 & 0.331 & 0.668 & 0.820 & 0.379 & 0.803 \\
 No NMS & 0.329 & 0.361 & 0.327 & 0.299 & 0.663 & 0.814 & 0.371 & 0.803 \\
 No MT & 0.351 & 0.402 & 0.339 & 0.312 & 0.757 & 0.918 & 0.502 & 0.852 \\
 \hline
 Constant Cov & 0.345 & 0.387 & 0.349 & 0.301 & 0.674 & 0.842 & 0.376 & 0.804 \\
 Trainable Cov & 0.355 & 0.406 & 0.356 & 0.304 & $\bm{0.635}$ & 0.770 & 0.365 & 0.770 \\
 Tr. Cov w Min & 0.359 & 0.405 & 0.349 & 0.323 & 0.647 & 0.791 & 0.376 & 0.773 \\
 
 \hline
 3Heads & $\bm{0.368}$ & 0.420 & 0.368 & 0.316 & 0.676 & 0.820 & 0.367 & 0.841 \\
\end{tabular}
\end{ruledtabular}
\end{table*}

\section{Evaluation}
\label{sec:results}

We evaluate the performance of Kraken on the Waymo Open Dataset and compare it to other entries on the
public leaderboards of the Waymo Prediction Challenge 2021 \cite{WaymoDataset}.

\subsection{Dataset}
\label{sec:Waymo-dataset}

Waymo dataset is generated from $103 354$ $20$-second segments that contain $10 \text{Hz}$ object tracks and map data of the area, including both stationary objects and traffic lights history.
These segments are cut into $9$-second fragments, in which the first second is interpreted as known history, while the remaining $8$ seconds are to be predicted by the model at $2\text{Hz}$ frequency. The prediction can include up to six different future realizations (modes) and a score (probability) should be assigned to each of them.
All agents are divided into three classes: (i) vehicles, including the flagged self-driving rover, (ii) pedestrians, and (iii) cyclists.
Some of the agents are marked as demonstrating interesting behavior (up to $8$ actors per scene) and the motion leaderboard metrics are calculated on this subset of actors.
Conveniently, Waymo dataset is divided into training, validation, and (hidden) testing parts. 

Waymo challenge consists of Motion Prediction and Interaction Prediction parts. In the former, each agent is treated separately and mode probabilities can have independent values for each agent. In the latter, one should predict a joint future for pairs of objects that are marked to demonstrate apparently interacting behavior. So, in this case modes probabilities should be shared between these two objects.

Waymo suggests a number of metrics to evaluate the submissions. 
All metrics are averaged over three prediction times ($3$, $5$, and $8$ seconds) and the three types of objects. The latter should make the model treat each actor type with equal importance despite the fact that on average there are much more vehicles in a typical scene than pedestrians and cyclists.
Considered metrics include a conventional distance-based minADE metrics (in its single and joint versions) and a diversity-inclined hit rate-based mAP metric \cite{WaymoDataset}, which is used for the leaderboard ranking.
A prediction is assumed to be a hit if it lies in the rectangular box centered by the ground truth and aligned along with it (longitudinal tolerance is twice as the perpendicular one).
The size of this box depends linearly on the timestamp and has an additional factor that takes into account agent's velocity at the scene time ($1 s$ from the beginning of the track).
Agents with a smaller initial velocity have a smaller hit window.
Another important aspect of this metric is that all agents are bucketed with respect to the motion direction (left, right, U-turn, etc.) and mAP is calculated inside each bucket separately by aggregating predictions from all scenes and evaluating the area under the precision-recall curve.
Finally, one should average the outcome among the buckets.

We evaluated our model on the public leaderboards of Waymo Prediction Challenge, which use hidden testing part of the Waymo dataset to evaluate the submissions. The top of the Motion prediction leaderboard as of October 2021 can be found in Tbl.~\ref{tbl:motion_leaderboard}. Our model scored second in this competition both with respect to mAP and minADE metrics.
In Tbl.~\ref{tbl:interaction_leaderboard} we show the head of the Interaction prediction leaderboard. Our model takes the first place with a significant margin in mAP compared to the subleading submissions.
With respect to minADE metrics, our model scores second after SceneTransformer on the Interaction leaderboard.

We complement evaluation on the Waymo Open Challenge by an ablation study, which is evaluated on the validation dataset  since the testing part is not freely available. 
We identify three most important mAP-boosting features as Mode Transformer block, Greedy Mode Processing (GMP), and tuning the model distribution \eqref{distribution_exponential} to Waymo hit window rules. Thus, we consider models, in which one of these features is excluded and present the resulting metrics in Tbl.~\ref{tbl:Ablations}. 
We consider several options for the non-Waymo tuned covariance matrix $\Sigma$ in \eqref{distribution_exponential}, which are explained in detail Sec.~\ref{sec:waymo_covariance}.
We conclude that out of these features, applying GMP algorithm improves mAP the most.
Mode Transformer block takes the second place with respect to mAP enhancement and provides the largest improvement for the minADE metric. 
We thus suggest that Mode Transformer can be used as a useful block in the future research on
motion prediction networks.

%%%%%%%

\section{Discussion and Conclusion}
\label{sec:conclusion}

In the current study, we presented Kraken, a motion prediction model capable of making multimodal future predictions and accounting for pairwise interaction of actors in a scene. 
In its backbone architecture, Kraken is a combination of modern neural network solutions including convolutional nets applied to the bird-eye-view feature maps, recursive neural nets encoding past trajectory of each actor, and several Transformer blocks enabling attention mechanism between the embeddings of actor, timestep, and mode of the future distribution. 
Such integrated solution allows to effectively process multiple inputs of different modalities, such as road situation, motion history of agents present in the scene, and traffic regulations. 

One of the main features of our work is the introduction of a novel Mode Transformer block  implementing attention mechanism between mode embeddings that encode components of the modeled future distribution. Compared to the baseline model, which samples future trajectories independently, Mode Transformer allows to cover the possible dynamics in the scene more effectively. In practice, it improves minADE metric on the Waymo dataset by up to $10\,cm$ for $8s$-long future trajectories.

Another crucial development of the current study is employment of a non-maximum suppression algorithm for a greedy processing of trajectory candidates. When applied to a pair of objects, it allows to transform factorized distribution for two actors into a correlated prediction representing physically plausible joint trajectories. Inclusion of this strategy resulted in a significant boost of the Interaction mAP metrics of the Waymo Interaction Prediction Challenge.

The results of our study suggest a number of open questions to be addressed in the future works. One of such questions is the legitimacy of excluding predictions with intersecting trajectories --- while collisions are rare, they do happen in road accidents and eliminating them completely may jeopardize the safety of the autonomous vehicle. Another possible direction of the future research is generalization of predicting pairwise interactions explored in the current work  to a full-fledged joint prediction for the entire scene or, alternatively, developing an algorithm for identifying interacting actors and reducing the problem to a collection of small-scale joint prediction instances.

As a confirmation of its effectiveness, Kraken held the first place on the public leaderboard of Waymo Interaction Prediction Challenge and the second place at the Waymo Motion Prediction Challenge when evaluated in October 2021.

\bibliography{kraken}% Produces the bibliography via BibTeX.

\end{document}